\documentclass{article} 
\usepackage{iclr2026_conference,times}
\usepackage{placeins}

\usepackage{amsmath,amsfonts,bm}

\def\eqref#1{equation~\ref{#1}}

\def\1{\bm{1}}

\DeclareMathAlphabet{\mathsfit}{\encodingdefault}{\sfdefault}{m}{sl}
\SetMathAlphabet{\mathsfit}{bold}{\encodingdefault}{\sfdefault}{bx}{n}

\usepackage{hyperref}
\usepackage{url}
\usepackage{graphicx}

\usepackage{amsmath}
\usepackage{amssymb}
\usepackage{algorithm}
\usepackage{algpseudocode}
\usepackage{booktabs}

\newcommand{\EIG}{\mathrm{EIG}}
\newcommand{\Hcal}{\mathcal{H}}
\newcommand{\Xcal}{\mathcal{X}}
\newcommand{\Dcal}{\mathcal{D}}
\newcommand{\LLM}{\mathrm{LLM}}
\newcommand{\MAP}{\mathrm{MAP}}
\newcommand{\PTS}{\mathrm{PTS}}

\title{Wild Guesses and Mild Guesses in Active Concept Learning}

\author{Anirudh Chari \\
Massachusetts Institute of Technology\\
Cambridge, MA, USA \\
\texttt{anichari@mit.edu} \\
\And
Neil Pattanaik \\
University of California, Berkeley \\
Berkeley, CA, USA \\
\texttt{neil\_pattanaik@berkeley.edu} \\
}

\iclrfinalcopy 
\begin{document}

\maketitle

\begin{abstract}
Human concept learning is typically active: learners choose which instances to query or test in order to reduce uncertainty about an underlying rule or category. Active concept learning must balance informativeness of queries against the stability of the learner that generates and scores hypotheses. We study this trade-off in a neuro-symbolic Bayesian learner whose hypotheses are executable programs proposed by a large language model (LLM) and reweighted by Bayesian updating. We compare a Rational Active Learner that selects queries to maximize approximate expected information gain (EIG) and the human-like Positive Test Strategy (PTS) that queries instances predicted to be positive under the current best hypothesis.
Across concept-learning tasks in the classic Number Game, EIG is effective when falsification is necessary (e.g., compound or exception-laden rules), but underperforms on simple concepts. We trace this failure to a support mismatch between the EIG policy and the LLM proposal distribution: highly diagnostic boundary queries drive the posterior toward regions where the generator produces invalid or overly specific programs, yielding a support-mismatch trap in the particle approximation.
PTS is information-suboptimal but tends to maintain proposal validity by selecting "safe" queries, leading to faster convergence on simple rules. Our results suggest that "confirmation bias" may not be a cognitive error, but rather a rational adaptation for maintaining tractable inference in the sparse, open-ended hypothesis spaces characteristic of human thought.
\end{abstract}

\section{Introduction}
Human concept learning is typically \emph{active}: learners select queries and interventions to reduce uncertainty about latent structure \citep{lake2017building}. We use active concept learning in reference to this query-driven setting when the goal is to identify a structured concept, such as a rule or program, rather than to optimize a predictive model over a fixed hypothesis class.
In Bayesian accounts such as Bayesian Program Learning (BPL), concepts correspond to hypotheses $h \in \Hcal$ (often programs), and learning is inference over $\Hcal$ conditioned on observed labels \citep{tenenbaum1999bayesian,tenenbaum2000rules,lake2015human}. A growing line of neuro-symbolic work replaces hand-engineered program proposals with large language models (LLMs) that generate executable code from natural-language prompts \citep{ellis2023human}. This substitution substantially expands the effective hypothesis space, but it also introduces a new bottleneck: inference and active learning are now limited by the support and calibration of the generator used to propose hypotheses. From a cognitive-science perspective, this ``proposal bottleneck'' operationalizes a concrete resource constraint on hypothesis generation, making it possible to test when human-like query heuristics are adaptive because they preserve tractable inference rather than maximize information per query.

This paper studies active Bayesian concept learning in an LLM-proposed program space. A natural baseline is to choose the next query by maximizing expected information gain (EIG), a normative criterion in optimal experiment design and active learning \citep{oaksford1994rational,nelson2005finding,settles2009active}. However, human learners frequently adopt a Positive Test Strategy (PTS), preferentially querying instances expected to be positive under their current hypothesis \citep{klayman1987confirmation}. PTS is often characterized as ``confirmation bias'' \citep{wason1960failure}, yet rational analyses show it can be adaptive under sparsity, cost, or asymmetric noise \citep{oaksford1994rational,navarro2011hypothesis}.

We find that in an LLM-driven neuro-symbolic learner, the EIG-optimal policy can be computationally adversarial even when it is information-theoretically sensible. On simple concepts, EIG repeatedly selects boundary cases that maximize predictive entropy under the current particle set. After observing the label, many particles are eliminated, and the learner must replenish with new hypotheses from the LLM. In precisely these high-diagnostic regimes, the proposal distribution often produces invalid, inconsistent, or overly specific programs, leading to particle degeneracy and slow recovery. We refer to this as a support-mismatch trap. In contrast, PTS tends to keep the learner within regions where the generator proposes stable, consistent hypotheses, trading off formal optimality for robust progress.

\paragraph{Contributions}
\begin{enumerate}
    \item We formalize an active learning loop for neuro-symbolic Bayesian concept learning with an LLM-based proposal distribution and particle posterior approximation.
    \item We identify and empirically characterize a failure mode where EIG-driven querying induces proposal-support collapse, degrading performance on simple concepts.
    \item We provide evidence that PTS can act as a stability-preserving heuristic in open-ended program spaces, clarifying one computational condition under which confirmation-style sampling is beneficial.
\end{enumerate}

\section{Problem Setup}
We study binary concept learning with active queries. An unknown target concept $h^\star \in \Hcal$ maps instances $x \in \Xcal$ to labels $y \in \{0,1\}$.
At time $t$, the learner has observed a dataset
\begin{equation}
\Dcal_t = \{(x_i,y_i)\}_{i=1}^t,
\end{equation}
and selects a new query $x_{t+1} \in \Xcal \setminus \{x_1,\ldots,x_t\}$, receiving $y_{t+1} = h^\star(x_{t+1})$.

In the Number Game experiments, $\Xcal = \{0,\dots,100\}$ and $\Hcal$ is an open-ended space of executable predicates (represented as Python functions) that may include arithmetic, modular, and digit-based properties. Because $\Hcal$ is effectively unbounded, exact posterior inference is intractable; we approximate $p(h \mid \Dcal_t)$ with a particle posterior induced by an LLM proposal distribution.

\section{Neuro-Symbolic Bayesian Learner}
We maintain a set of $N$ hypothesis particles $\{h_i\}_{i=1}^N$ with weights $\{w_i\}_{i=1}^N$ approximating the posterior:
\begin{equation}
p(h \mid \Dcal_t) \approx \sum_{i=1}^N w_i \, \delta(h=h_i), \qquad \sum_{i=1}^N w_i = 1.
\end{equation}
Following \citet{ellis2023human}, we use an LLM as a proposal distribution $q_{\LLM}(h \mid \Dcal_t)$ by prompting it with the current observations and requesting candidate executable programs.
We filter proposals for syntactic validity and consistency with $\Dcal_t$; invalid or inconsistent programs are discarded.

After observing $(x_{t+1},y_{t+1})$, we update each weight by checking consistency with the new label. When particle diversity collapses (effective sample size below a threshold), we rejuvenate by drawing additional proposals from $q_{\LLM}(h \mid \Dcal_{t+1})$ and reweighting. In practice, rejuvenation quality depends strongly on whether the new dataset $\Dcal_{t+1}$ lies within the proposal distribution's reliable support.

For each particle $h_i$, we define an unnormalized weight
\begin{equation}
\tilde{w}_i \propto p(\Dcal_t \mid h_i)\, p(h_i),
\end{equation}
and normalize across particles.

\paragraph{Likelihood (size principle)}
We use the size principle for positive examples:
\begin{equation}
p(\Dcal_t \mid h) \propto \left(\frac{1}{|h|}\right)^{|\Dcal_t^+|},
\end{equation}
where $|h|$ denotes the extension size (the number of $x \in \Xcal$ such that $h(x)=1$) and $\Dcal_t^+$ is the set of observed positive examples. Intuitively, this penalizes overly broad concepts.

\paragraph{Prior (simplicity bias)}
We include a code-simplicity prior based on model scoring of the program text (e.g., average token log-probability under a code model), which biases toward shorter and more likely programs:
\begin{equation}
p(h) \propto \exp\!\left(\frac{1}{|h|_{\text{tok}}}\sum_{j=1}^{|h|_{\text{tok}}}\log p_{\text{Code}}(t_j \mid t_{<j})\right).
\end{equation}
This prior is intended as a computational proxy for description-length biases used in program induction.

\section{Active Query Policies}
At each step, a policy selects the next query $x \in \Xcal \setminus \{x_1,\ldots,x_t\}$. We use a passive policy baseline, which samples $x$ uniformly at random from unqueried instances.

\subsection{Rational Active Learner: approximate EIG}
Given the particle posterior, the predictive probability of label $y$ at instance $x$ is
\begin{equation}
p(y \mid x,\Dcal_t) \approx \sum_{i=1}^N w_i \, \mathbb{I}[h_i(x)=y].
\end{equation}
We approximate the expected information gain of querying $x$ as the expected reduction in posterior entropy:
\begin{equation}
\EIG(x;\Dcal_t)
= H\!\left(p(h \mid \Dcal_t)\right)
- \sum_{y \in \{0,1\}} p(y \mid x,\Dcal_t)\, H\!\left(p(h \mid \Dcal_t \cup \{(x,y)\})\right),
\end{equation}
where $H(\cdot)$ is Shannon entropy computed over the discrete particle weights. The EIG policy selects
\begin{equation}
x^{\star}_{\EIG} = \arg\max_{x \in \Xcal \setminus \{x_1,\ldots,x_t\}} \EIG(x;\Dcal_t).
\end{equation}
This criterion is normative with respect to the current posterior approximation; it does not account for failures of the hypothesis generator after updating.

\subsection{Positive Test Strategy (PTS)}
PTS queries instances predicted to be positive under the current MAP hypothesis:
\begin{equation}
h_{\MAP} = \arg\max_{h_i} w_i, \qquad
x^{\star}_{\PTS} \sim \mathrm{Unif}\Big(\{x \in \Xcal \setminus \{x_1,\ldots,x_t\} : h_{\MAP}(x)=1\}\Big).
\end{equation}
Unlike EIG, PTS is not designed to maximize disambiguation. Its computational advantage is that it tends to avoid queries that sharply constrain the hypothesis space in ways that exceed the generator's ability to propose alternatives.

\begin{algorithm}[t]
\caption{Active neuro-symbolic Bayesian concept learning (particle approximation).}
\label{alg:active_bpl}
\begin{algorithmic}[1]
\Require Instance space $\Xcal$, budget $T$, particles $N$, policy $\pi \in \{\EIG,\PTS,\mathrm{rand}\}$
\State Initialize $\Dcal_0 \gets \emptyset$
\State Propose initial particles $\{h_i\}_{i=1}^N \sim q_{\LLM}(h \mid \Dcal_0)$; set weights $w_i \propto p(\Dcal_0 \mid h_i)p(h_i)$
\For{$t = 0,1,\dots,T-1$}
    \State Select query $x_{t+1} \gets \pi(\Xcal,\Dcal_t,\{h_i,w_i\}_{i=1}^N)$
    \State Observe label $y_{t+1} \gets h^\star(x_{t+1})$
    \State $\Dcal_{t+1} \gets \Dcal_t \cup \{(x_{t+1},y_{t+1})\}$
    \State Update weights: $w_i \gets w_i \cdot \mathbb{I}[h_i(x_{t+1})=y_{t+1}]$; normalize
    \If{particle degeneracy (low ESS)}
        \State Propose additional hypotheses from $q_{\LLM}(h \mid \Dcal_{t+1})$, filter for consistency
        \State Reweight all particles by $p(\Dcal_{t+1}\mid h)p(h)$; normalize
    \EndIf
\EndFor
\end{algorithmic}
\end{algorithm}

\section{Results}
We evaluate on the Number Game with $\Xcal=\{0,\dots,100\}$. Target concepts are grouped into Easy (single salient property), Medium (boolean combinations or constraints), and Hard (non-local or idiosyncratic structure). At each iteration we prompt an LLM to propose candidate Python predicates consistent with the current dataset $\Dcal_t$. We use Gemini 2.5 Flash as the proposal engine. Proposals are filtered for syntactic validity and consistency with all observed labels.

Each trial runs for at most $T=50$ queries. We declare convergence when the particle posterior assigns confidence at least $0.95$ to the MAP hypothesis (within the maintained particle set). Runs that do not converge within the budget are marked \textbf{DNF} (did not finish). For each target concept and strategy, we report the number of queries to convergence (lower is better). Because the posterior is approximate, this metric captures both informativeness of queries and robustness of the proposal-and-filtering pipeline.

Table~\ref{tab:results} summarizes query counts across strategies.

\begin{table*}[t]
\centering
\caption{Number of queries to convergence ($\leq 50$) in the Number Game. \textbf{DNF} indicates no convergence within the budget.}
\begin{tabular*}{\textwidth}{@{\extracolsep{\fill}} lccc}
\toprule
\textbf{Rule} & \textbf{EIG} & \textbf{PTS} & \textbf{Passive} \\
\midrule
\multicolumn{4}{l}{\textbf{Easy rules}} \\
\midrule
Square numbers            & 3  & 2   & 2  \\
Multiples of 4            & 7  & 4   & 2  \\
Odd numbers               & 8  & 6   & 5  \\
Powers of 2               & 5  & 2   & 2  \\
\midrule
\multicolumn{4}{l}{\textbf{Medium rules}} \\
\midrule
Even numbers $< 30$        & 17 & 31  & 17 \\
Multiples of 3 or 7        & 8  & \textbf{DNF} & 13 \\
Odd multiples of 3         & 14 & 35  & 21 \\
Numbers ending in 6        & 4  & 7   & 7  \\
\midrule
\multicolumn{4}{l}{\textbf{Hard rules}} \\
\midrule
Digits sum to $< 8$        & 43 & \textbf{DNF} & \textbf{DNF} \\
Remainder $=5 \pmod{9}$    & 31 & \textbf{DNF} & 41  \\
One less than a prime      & \textbf{DNF} & \textbf{DNF} & \textbf{DNF} \\
Twice a square minus 2     & \textbf{DNF} & \textbf{DNF} & \textbf{DNF} \\
\bottomrule
\end{tabular*}
\label{tab:results}
\end{table*}

\paragraph{Overthinking to Underperform}
On Easy rules, EIG is not reliably better than passive sampling and is often worse than PTS. Qualitatively, EIG tends to select boundary points that maximize predictive entropy under the current particle set. After observing the label, the posterior can collapse onto a narrow region where few valid hypotheses remain under the LLM proposal distribution. The subsequent rejuvenation step then produces many invalid or overly specific programs, delaying recovery. This is the \textbf{support-mismatch trap}: the policy is ``optimal'' for the current approximation but induces datasets for which the generator has poor coverage.

\paragraph{Falsification is Useful}
For Medium rules, the pattern reverses: PTS frequently fails to seek counterevidence needed to reject plausible subsets (e.g., committing to a stricter divisor rule when the target is looser). EIG more consistently identifies disambiguating queries, reducing time to convergence on several compound concepts.

\paragraph{Generator Bottleneck}
All methods struggle on Hard rules, with EIG sometimes succeeding where others do not. However, the overall failure rate suggests that query selection cannot compensate for a proposal distribution that assigns negligible mass to the target concept. Active learning can amplify existing modeling capacity, but it cannot reliably recover hypotheses that the generator almost never proposes.

\section{Analysis and Discussion}
\subsection{A computational account of confirmation-style sampling}
PTS improves performance on simple concepts not by maximizing information per query, but by maintaining a dataset that stays within the generator's reliable regime. In an LLM-driven learner, ``rationality'' must be evaluated with respect to the full system: inference and hypothesis proposal. Under proposal-support limitations, PTS can be viewed as a stability-preserving heuristic that avoids high-risk queries that cause particle collapse and low-quality regeneration.

\subsection{When should an agent switch strategies?}
The results are consistent with a meta-policy hypothesis: use low-cost, stability-preserving sampling early (PTS-like) and shift toward falsification-heavy sampling (EIG-like) when progress stalls, contradictions appear, or concepts appear compositional. Testing such a switch requires explicit criteria (e.g., posterior stagnation, effective sample size, or repeated generator failure) and is a concrete direction for workshop-relevant human-inspired AI reasoning.

\subsection{Implications for neurosymbolic systems}
The support-mismatch trap suggests that active learning objectives should incorporate generator awareness. Two practical directions are: generator-regularized acquisition functions that penalize queries expected to induce low proposal success, and robust rejuvenation mechanisms (e.g., structured DSL backstops or constrained decoding) that maintain coverage under sharper constraints. We do not explore these here; the present goal is to isolate and document the failure mode. 

From a cognitive perspective, the same analysis offers a functional interpretation of PTS. If human hypothesis generation is similarly sparse and support-limited, then PTS can be viewed as a heuristic for keeping inference in a ``high-coverage'' region of hypothesis space. In this view, preferentially sampling predicted positives is not primarily about seeking confirmatory evidence per se, but about avoiding queries that would force abrupt shifts into regions where few coherent hypotheses can be generated or maintained. This provides a concrete mechanism by which PTS can enable rapid convergence on simple concepts: it preserves stable hypothesis proposal and prevents the kind of regeneration failures that arise after highly diagnostic boundary queries.

\section{Conclusion}
We studied active concept learning in a neuro-symbolic Bayesian learner whose hypotheses are programs proposed by an LLM. Although expected information gain is a principled acquisition objective, we find that it can be counterproductive in open-ended program spaces because it induces datasets that fall outside the generator's reliable support, causing particle degeneracy and slow recovery. A Positive Test Strategy, while not information-optimal, can act as a stability-preserving heuristic that accelerates learning on simple concepts. These findings motivate generator-aware active learning objectives and meta-policies that adaptively trade off falsification against stability---a direct bridge between cognitive regularities in human inquiry and practical design constraints in modern neurosymbolic AI.

\bibliography{iclr2026_conference}
\bibliographystyle{iclr2026_conference}

\newpage
\appendix
\section{Additional Results and Trajectories}
\FloatBarrier

\begin{figure*}[t]
    \centering
    \includegraphics[width=\linewidth]{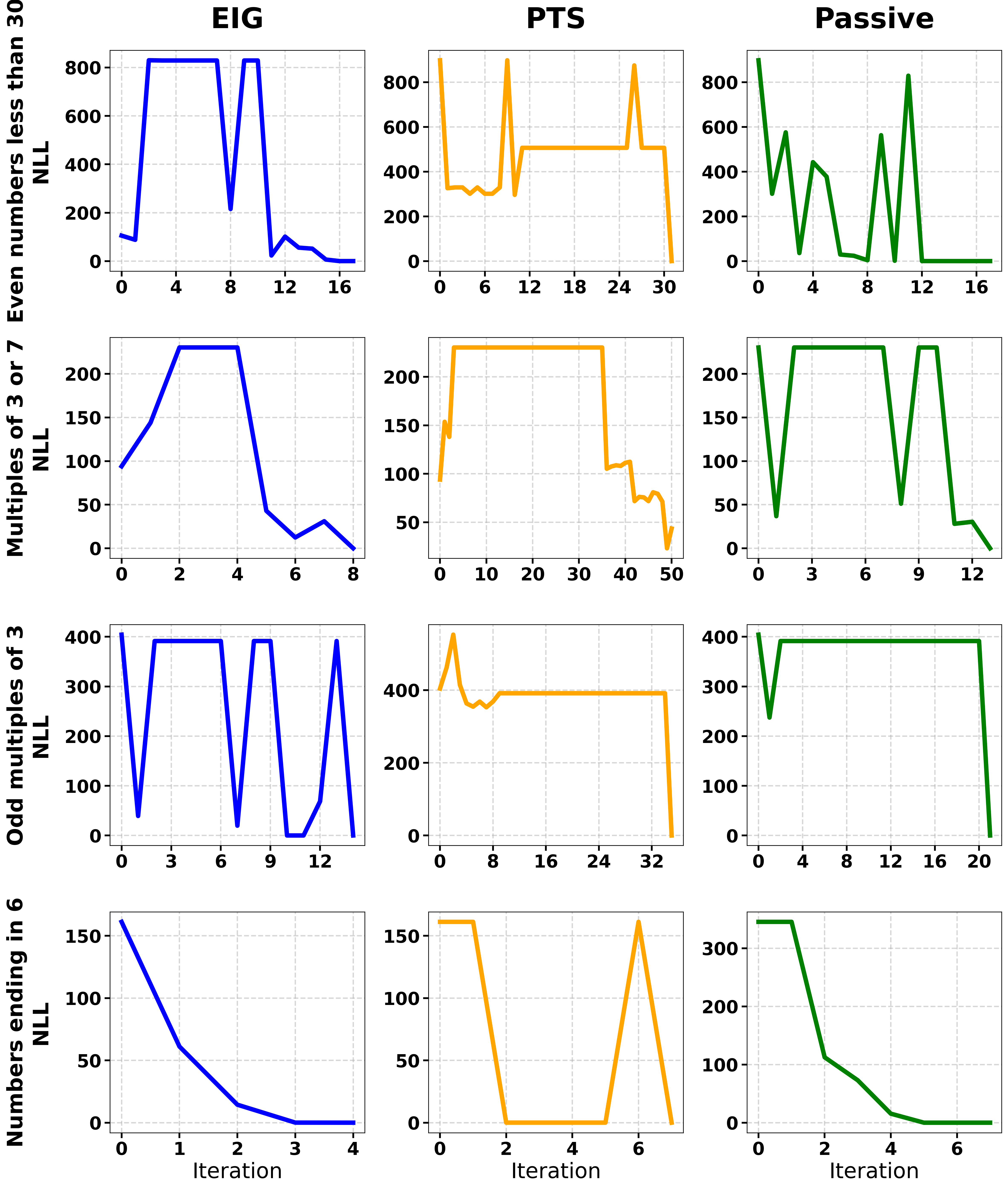}
    \caption{Negative log-likelihood (NLL) of the true concept over time for representative Medium-rule trials (lower is better).}
    \label{fig:nll_plots}
\end{figure*}

\begin{table*}[h]
\centering
\small
\renewcommand{\arraystretch}{1.1}
\begin{tabular*}{\textwidth}{@{\extracolsep{\fill}} c ll ll}
\toprule
& \multicolumn{2}{c}{\textbf{Rational Active Learner (EIG)}}
& \multicolumn{2}{c}{\textbf{Positive Test Strategy (PTS)}} \\
\cmidrule(lr){2-3} \cmidrule(lr){4-5}
\textbf{Iter}
& \textbf{Top hypothesis} ($P_{\text{conf}}$)
& \textbf{Query}
& \textbf{Top hypothesis} ($P_{\text{conf}}$)
& \textbf{Query} \\
\midrule
\multicolumn{5}{l}{\textit{\textbf{(a) Target concept: Multiples of 4}}} \\
\midrule
1 & Divisible by 4 ($0.83$) & \textbf{64} (Yes)
  & Power of 2 ($0.72$) & \textbf{1} (No) \\
2 & Power of 2 ($0.94$) & \textbf{32} (Yes)
  & Divisible by 4 ($0.99$) & \textbf{36} (Yes) \\
3 & Power of 2 ($1.00$) & \textbf{22} (No)
  & Divisible by 4 ($0.99$) & \textbf{24} (Yes) \\
4 & Power of 2 ($1.00$) & \textbf{77} (No)
  & Divisible by 4 ($1.00$) & \textit{-- Converged --} \\
5 & Power of 2 ($1.00$) & \textbf{36} (Yes) & & \\
6 & Divisible by 4 ($0.97$) & \textbf{81} (No) & & \\
7 & Divisible by 4 ($1.00$) & \textit{-- Converged --} & & \\
\midrule
\multicolumn{5}{l}{\textit{\textbf{(b) Target concept: Powers of 2}}} \\
\midrule
1 & Power of 2 ($0.59$) & \textbf{0} (No)
  & Power of 2 ($0.59$) & \textbf{64} (Yes) \\
2 & Power of 2 ($0.72$) & \textbf{9} (No)
  & Power of 2 ($1.00$) & \textit{-- Converged --} \\
3 & Power of 2 ($0.99$) & \textbf{1} (Yes) & & \\
4 & Power of 2 ($0.96$) & \textbf{3} (No) & & \\
\bottomrule
\end{tabular*}
\caption{Example learning trajectories under EIG and PTS.}
\label{tab:trajectories}
\end{table*}

\end{document}